\documentclass[10pt,twocolumn]{article}
\usepackage[utf8]{inputenc}
\usepackage[T1]{fontenc}
\usepackage{amsmath,amssymb}
\usepackage{graphicx}
\usepackage{booktabs}
\usepackage{hyperref}
\usepackage[margin=0.75in]{geometry}
\usepackage{caption}
\usepackage{subcaption}
\usepackage[numbers]{natbib}
\usepackage{xcolor}
\usepackage{enumitem}
\usepackage{float}
\usepackage{multirow}
\usepackage{tabularx}
\usepackage{tcolorbox}
\usepackage{titlesec}

\definecolor{darkblue}{RGB}{27,58,92}
\definecolor{medblue}{RGB}{46,117,182}
\definecolor{accentblue}{RGB}{46,117,182}
\definecolor{lightgray}{RGB}{245,245,245}
\definecolor{tableheader}{RGB}{220,230,241}
\hypersetup{colorlinks=true, linkcolor=darkblue, citecolor=darkblue, urlcolor=darkblue}

\titleformat{\section}
    {\normalfont\large\bfseries\color{darkblue}}
    {\thesection}{0.5em}{}
    [\vspace{0.1em}\titlerule]

\titleformat{\subsection}
    {\normalfont\normalsize\bfseries\color{medblue}}
    {\thesubsection}{0.5em}{}

\titlespacing*{\section}{0pt}{14pt plus 2pt minus 2pt}{6pt plus 1pt}
\titlespacing*{\subsection}{0pt}{10pt plus 2pt minus 1pt}{4pt plus 1pt}

\title{%
\vspace{-0.3cm}%
\rule{\textwidth}{2.5pt}\\[0.35cm]%
{\fontsize{16}{19}\selectfont\textbf{The AI Skills Shift: Mapping Skill Obsolescence,\\[0.15cm]
Emergence, and Transition Pathways in the LLM Era}}\\[0.25cm]%
\rule{\textwidth}{0.6pt}%
\vspace{0.1cm}%
}

\author{
\textbf{Rudra Jadhav} \\[0.1cm]
{\small Department of Computer Science}\\
{\small Savitribai Phule Pune University}\\
{\small Pune, India}\\
{\small\texttt{roodra.jadhav@gmail.com}}
\and
\textbf{Janhavi Danve} \\[0.1cm]
{\small Department of Computer Science}\\
{\small Savitribai Phule Pune University}\\
{\small Pune, India}\\
{\small\texttt{janhavi.danve@gmail.com}}
}

\date{\small\textcolor{gray}{April 2026 \quad$\vert$\quad Preprint --- Under Review}}

\begin{document}
\maketitle
\thispagestyle{empty}

% ================================================================
\begin{tcolorbox}[colback=lightgray, colframe=accentblue, boxrule=0.5pt, arc=2mm, left=6pt, right=6pt, top=4pt, bottom=4pt]
\small
\textbf{Abstract.} As Large Language Models reshape the global labor market, policymakers and workers need empirical data on which occupational skills may be most susceptible to automation. We present the \textbf{Skill Automation Feasibility Index (SAFI)}, benchmarking four frontier LLMs---LLaMA 3.3 70B, Mistral Large, Qwen 2.5 72B, and Gemini 2.5 Flash---across \textbf{263 text-based tasks} spanning all \textbf{35 skills} in the U.S. Department of Labor's O*NET taxonomy (\textbf{1,052 total model calls, 0\% failure rate}). Cross-referencing with real-world AI adoption data from the Anthropic Economic Index (756 occupations, 17,998 tasks), we propose an \textbf{AI Impact Matrix}---an interpretive framework that positions skills along four quadrants: High Displacement Risk, Upskilling Required, AI-Augmented, and Lower Displacement Risk. Key findings: \textbf{(1)} Mathematics (SAFI: 73.2) and Programming (71.8) receive the highest automation feasibility scores; Active Listening (42.2) and Reading Comprehension (45.5) receive the lowest; \textbf{(2)} a ``capability-demand inversion'' where skills most demanded in AI-exposed jobs are those LLMs perform \textit{least well} at in our benchmark; \textbf{(3)} 78.7\% of observed AI interactions are augmentation, not automation; \textbf{(4)} all four models converge to similar skill profiles (3.6-point spread), suggesting that text-based automation feasibility may be more skill-dependent than model-dependent. SAFI measures LLM performance on text-based representations of skills, not full occupational execution. All data, code, and model responses are open-sourced.

\end{tcolorbox}

% ================================================================
\section{Introduction}

The rapid advancement of Large Language Models has shifted the conversation about AI's economic impact from speculative forecasting to observable reality. In February 2026, JPMorgan Chase CEO Jamie Dimon confirmed that his bank has already experienced AI-driven workforce displacement, stating: ``We have displaced people from AI, and we offer them other jobs'' \cite{dimon2026feb}. Speaking at the Hill \& Valley Forum in March 2026, he warned that AI-driven disruption ``may be quicker'' than past technological transitions and called for coordinated government-business efforts to ``retrain, reskill, and redeploy'' affected workers \cite{dimon2026march}. Goldman Sachs CEO David Solomon offered a contrasting perspective, declaring ``I'm not in the job apocalypse camp'' while acknowledging that the pace of AI adoption means ``the short-term disruption might be a little bit higher'' than prior technology shifts \cite{solomon2026}. Anthropic CEO Dario Amodei issued perhaps the starkest warning, predicting AI could eliminate up to 50\% of entry-level white-collar jobs within five years \cite{amodei2026}.

These industry perspectives are supported by empirical data. Anthropic's Economic Index, based on analysis of over four million Claude conversations, found that AI usage primarily concentrates in software development and writing tasks, with approximately 36\% of occupations using AI for at least a quarter of their associated tasks \cite{handa2025anthropic}. The ``Agents of Chaos'' study \cite{shapira2026agentschaos} demonstrated through live deployment that autonomous AI agents exhibit both significant capabilities and critical failure modes---underscoring the urgency of systematic skill-level evaluation.

Yet a critical gap persists: while we know AI is being adopted, relatively few studies offer empirical data on \textit{which specific skills} LLMs can perform in text-based settings and how this capability relates to the labor market's skill structure. Existing studies rely on expert opinion \cite{eloundou2023gpts}, employer surveys \cite{wef2025}, or theoretical exposure indices \cite{pew2023}---few directly benchmark LLM capabilities against the standardized skill taxonomy that underpins workforce planning.

This paper addresses that gap with three contributions:

\begin{enumerate}[leftmargin=*, itemsep=2pt, topsep=4pt]
\item \textbf{The Skill Automation Feasibility Index (SAFI):} An empirically-derived score (0--100) for each of the 35 O*NET skills, based on benchmarking four frontier LLMs across 263 purpose-designed tasks with 1,052 total responses.

\item \textbf{The AI Impact Matrix:} An interpretive framework that cross-references SAFI scores with real-world AI adoption data from the Anthropic Economic Index, positioning skills along four quadrants to inform workforce planning discussions.

\item \textbf{The Capability-Demand Inversion:} Evidence from our benchmark that skills most demanded in AI-exposed occupations are those LLMs score \textit{lowest} on in text-based evaluation---consistent with the interpretation that the current AI wave is augmentation-dominant rather than automation-dominant.
\end{enumerate}

% ================================================================
\section{Related Work}

\subsection{AI and Labor Market Impact}
The study of technology's impact on employment has a rich history \cite{autor2015automation, acemoglu2020robots}. \cite{eloundou2023gpts} estimated that approximately 80\% of the U.S. workforce could have at least 10\% of their tasks affected by GPTs, based on expert annotations of O*NET tasks. However, their methodology relied on subjective human ratings rather than empirical capability testing. The Anthropic Economic Index \cite{handa2025anthropic, appel2025geographic} represents the most comprehensive empirical study of actual AI usage patterns, analyzing millions of conversations mapped to O*NET occupations and tasks. Their findings established that augmentation dominates automation by roughly 4:1 in current usage patterns. More recently, Anthropic's economic primitives framework \cite{appel2026primitives} introduced measures of task complexity, skill level, and AI autonomy, finding that more complex tasks are actually sped up \textit{more} by AI---a finding our results complement at the skill level.

\subsection{LLM Evaluation and Benchmarking}
Standard LLM benchmarks (MMLU, HumanEval, MATH) measure academic performance but do not map onto workforce skills \cite{hendrycks2021mmlu, chen2021humaneval}. The ``Agents of Chaos'' study \cite{shapira2026agentschaos} deployed six autonomous agents with persistent memory, email, and shell access, documenting eleven failure cases alongside six genuine safety behaviors---reinforcing the need for skill-level evaluation in real-world contexts. Additionally, recent work on LLM evaluation biases has shown that model outputs are influenced by surface-level features such as writing style and linguistic register \cite{jadhav2026grading}, a finding that motivated our choice of heuristic scoring over LLM-as-judge approaches in this study.

\subsection{Workforce Transition Frameworks}
The World Economic Forum \cite{wef2025} projects skill demand based on employer surveys across 55 economies. Pew Research \cite{pew2023} classified occupations by AI exposure using O*NET work activities. Our work differs from both by: (a) operating at the \textit{skill} level rather than occupation or activity level; (b) using empirical LLM benchmarks rather than surveys; and (c) cross-referencing with actual AI adoption data from production systems.

% ================================================================
\section{Data and Methods}

\subsection{Data Sources}

\textbf{O*NET Database (v30.2).} The U.S. Department of Labor's Occupational Information Network provides importance (IM) and level (LV) ratings for 35 skills across 1,016 occupations. Skills are organized into seven categories: Content (6 skills), Process (4), Social (6), Complex Problem Solving (1), Technical (11), Systems (3), and Resource Management (4).

\textbf{Anthropic Economic Index.} We utilize data from five releases (February 2025 through March 2026): job exposure scores for 756 occupations (range: 0.0 to 0.745), task penetration rates for 17,998 O*NET tasks, and automation versus augmentation interaction patterns for 3,364 tasks with five interaction modes (directive, feedback loop, task iteration, validation, learning).

\textbf{LLM Benchmark Data (ours).} 263 tasks across all 35 O*NET skills administered to four frontier LLMs, yielding 1,052 responses with a 100\% completion rate.

\subsection{Skill Taxonomy}

\begin{figure}[h!]
\centering
\includegraphics[width=\linewidth]{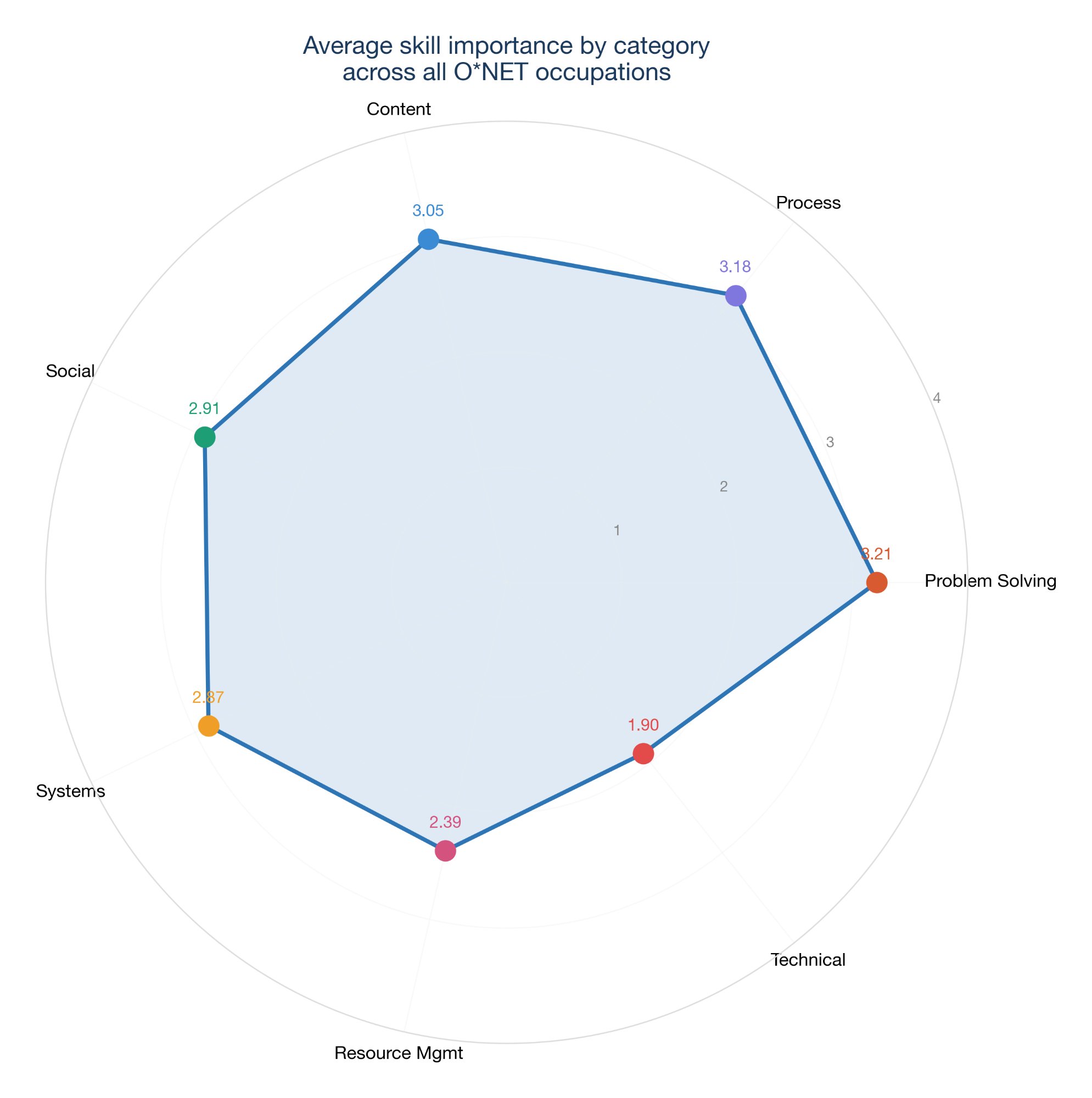}
\caption{Average skill importance by category across all 1,016 O*NET occupations. Process and Content skills are universally important; Technical skills are more specialized.}
\label{fig:radar}
\end{figure}

Figure~\ref{fig:radar} shows the average importance of each skill category across all O*NET occupations. Process Skills (3.18) and Content Skills (3.05) rank highest, reflecting their universal importance across the labor market. Technical Skills (1.90) rank lowest because the category includes highly specialized skills (Installation: 1.24, Equipment Maintenance: 1.71) that apply to a narrow set of occupations.

\subsection{AI Exposure Landscape}

\begin{figure}[h!]
\centering
\includegraphics[width=\linewidth]{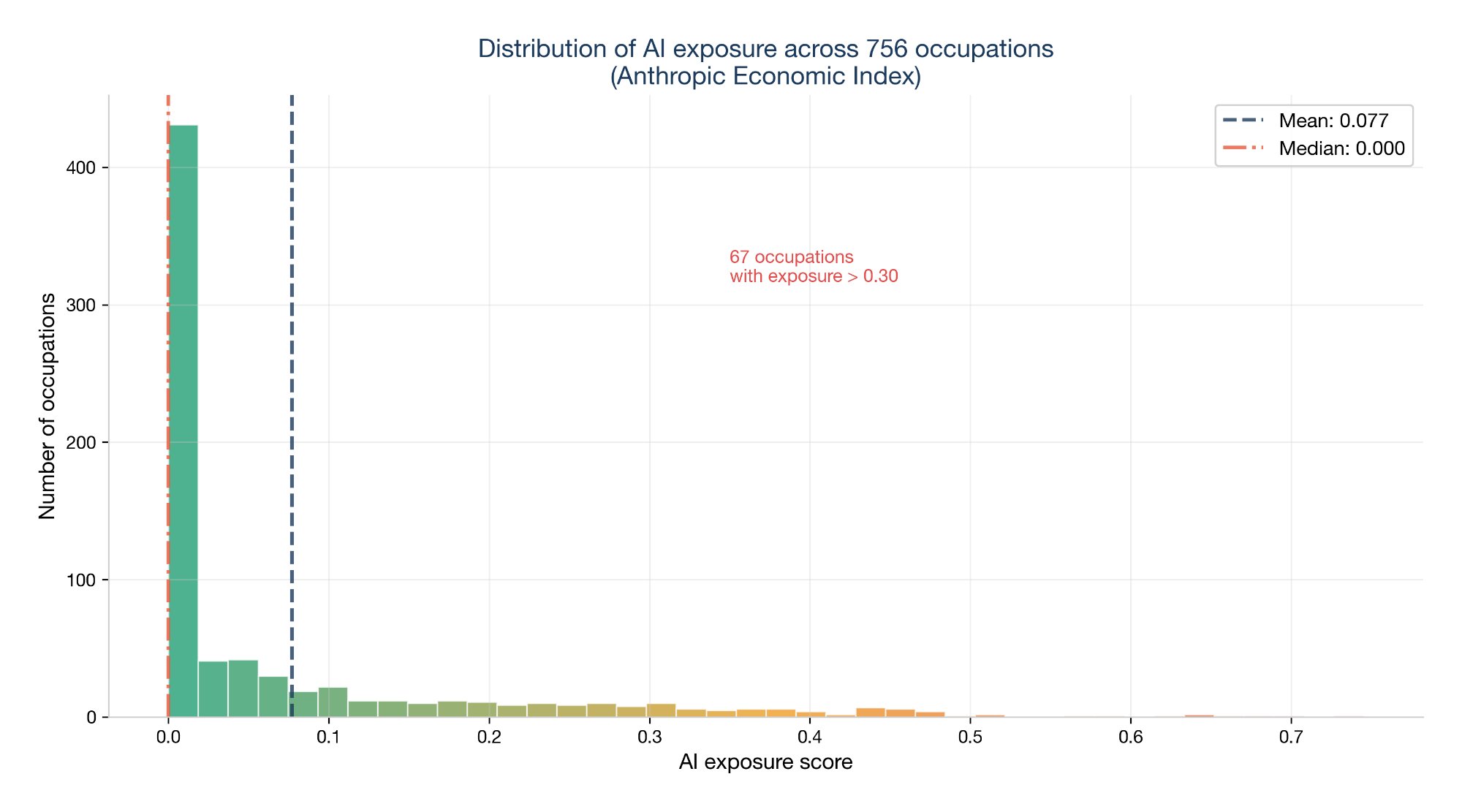}
\caption{Distribution of AI exposure across 756 occupations from the Anthropic Economic Index. Most occupations have minimal AI exposure; a long tail of 67 occupations exceeds 0.30.}
\label{fig:exposure_dist}
\end{figure}

The distribution of AI exposure across 756 occupations is heavily right-skewed (Figure~\ref{fig:exposure_dist}), with a mean exposure of 0.077 and median of 0.000. Only 67 occupations (8.9\%) have exposure scores above 0.30. The top five most AI-exposed occupations are Computer Programmers (0.745), Customer Service Representatives (0.701), Data Entry Keyers (0.671), Medical Records Specialists (0.667), and Market Research Analysts (0.648).

\begin{figure}[h!]
\centering
\includegraphics[width=\linewidth]{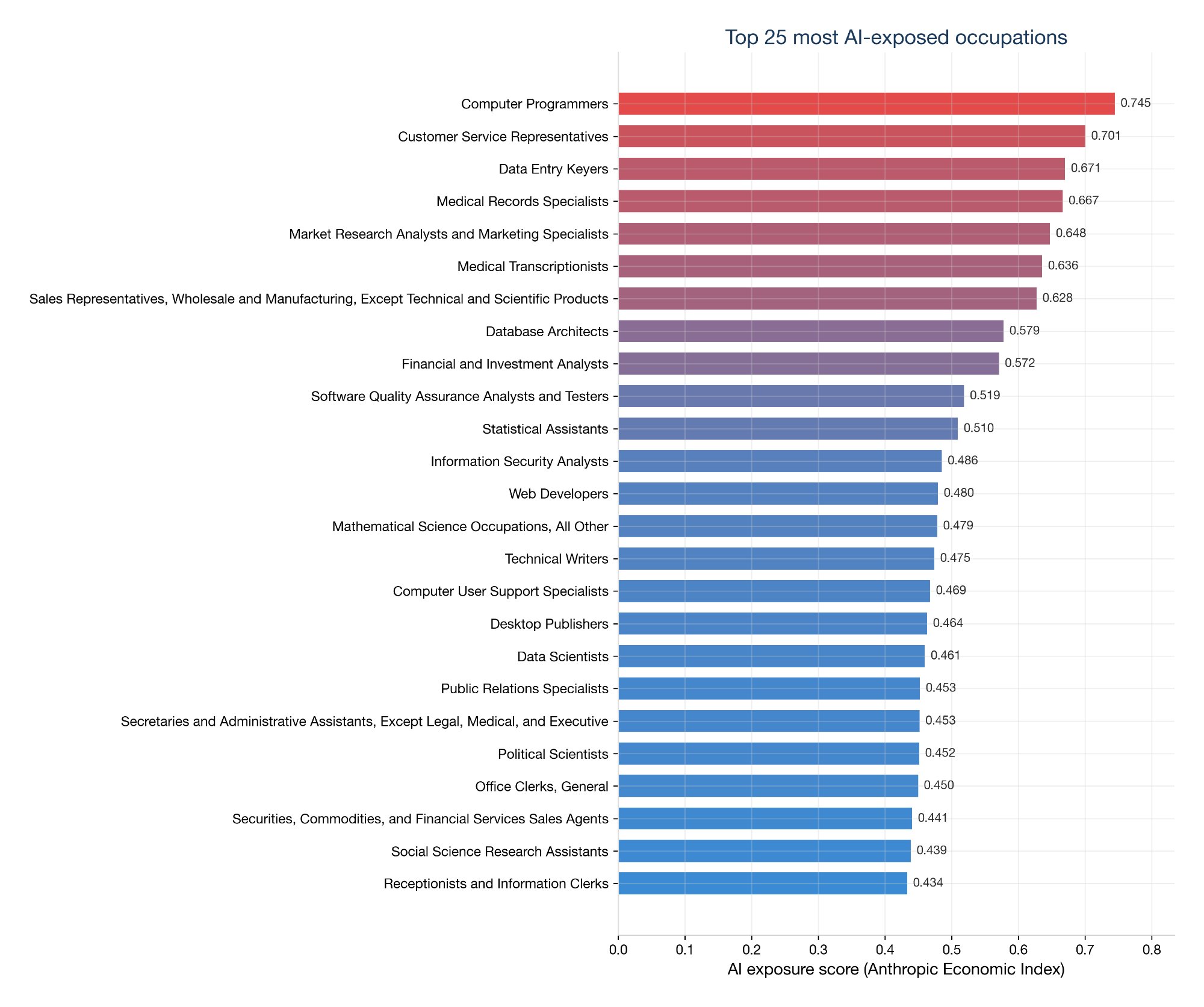}
\caption{The 25 most AI-exposed occupations. Computing, data, and customer-facing roles dominate.}
\label{fig:top25}
\end{figure}

\subsection{LLM Benchmark Design}

\subsubsection{Task Battery}
We designed 263 tasks covering all 35 O*NET skills at three difficulty levels (easy, medium, hard). Tasks are text-based prompts designed to elicit the \textit{cognitive and communicative dimensions} of each skill as defined by O*NET; they do not capture physical, embodied, or real-time interactive aspects of skill execution. For example, Reading Comprehension tasks require identifying contradictions between passages and evaluating methodological limitations; Negotiation tasks present multi-party scenarios requiring strategic thinking; Programming tasks require writing functional code and debugging errors. Each task includes a standardized grading rubric.

\subsubsection{Model Selection}
Four LLMs were selected for geographic, architectural, and licensing diversity:
\begin{itemize}[leftmargin=*, itemsep=1pt, topsep=2pt]
\item \textbf{LLaMA 3.3 70B} (Meta, US, open-source) via Groq
\item \textbf{Mistral Large} (Mistral AI, France, closed-source)
\item \textbf{Qwen 2.5 72B} (Alibaba, China, open-source) via HuggingFace
\item \textbf{Gemini 2.5 Flash} (Google, US, closed-source)
\end{itemize}

All models received identical prompts with temperature 0.3. The pipeline achieved \textbf{1,052 successful responses with zero failures}, executed over approximately 172 minutes with built-in resume capability and rate limiting.

\subsubsection{Scoring Methodology}
To avoid LLM-as-judge bias (where a model evaluates its own output favorably), responses were scored using a multi-signal heuristic engine across four dimensions: Response Completeness (0--3), Response Depth (0--3), Reasoning Quality (0--2), and Difficulty-Adjusted Bonus (0--2). Skill-specific adjustments add credit for mathematical calculations in Mathematics tasks and functional code in Programming tasks. Total scores range 0--10.

\subsubsection{SAFI Computation}
The Skill Automation Feasibility Index is computed as the average normalized score across all models and tasks for a given skill:
\begin{equation}
\text{SAFI}(s) = \frac{100}{|M|} \sum_{m \in M} \frac{1}{|T_s|} \sum_{t \in T_s} \frac{\text{score}(t, m)}{10}
\end{equation}
where $s$ is a skill, $M$ is the set of 4 models, and $T_s$ is the set of tasks for skill $s$. SAFI ranges from 0 (no measurable text-based performance) to 100 (perfect performance on all text-based tasks). Importantly, SAFI reflects LLM performance on textual representations of skills---it does not directly measure the feasibility of automating the full occupational contexts in which these skills are applied.

% ================================================================
\section{Results}

\subsection{SAFI Scores Across 35 Skills}

\begin{figure}[h!]
\centering
\includegraphics[width=\linewidth]{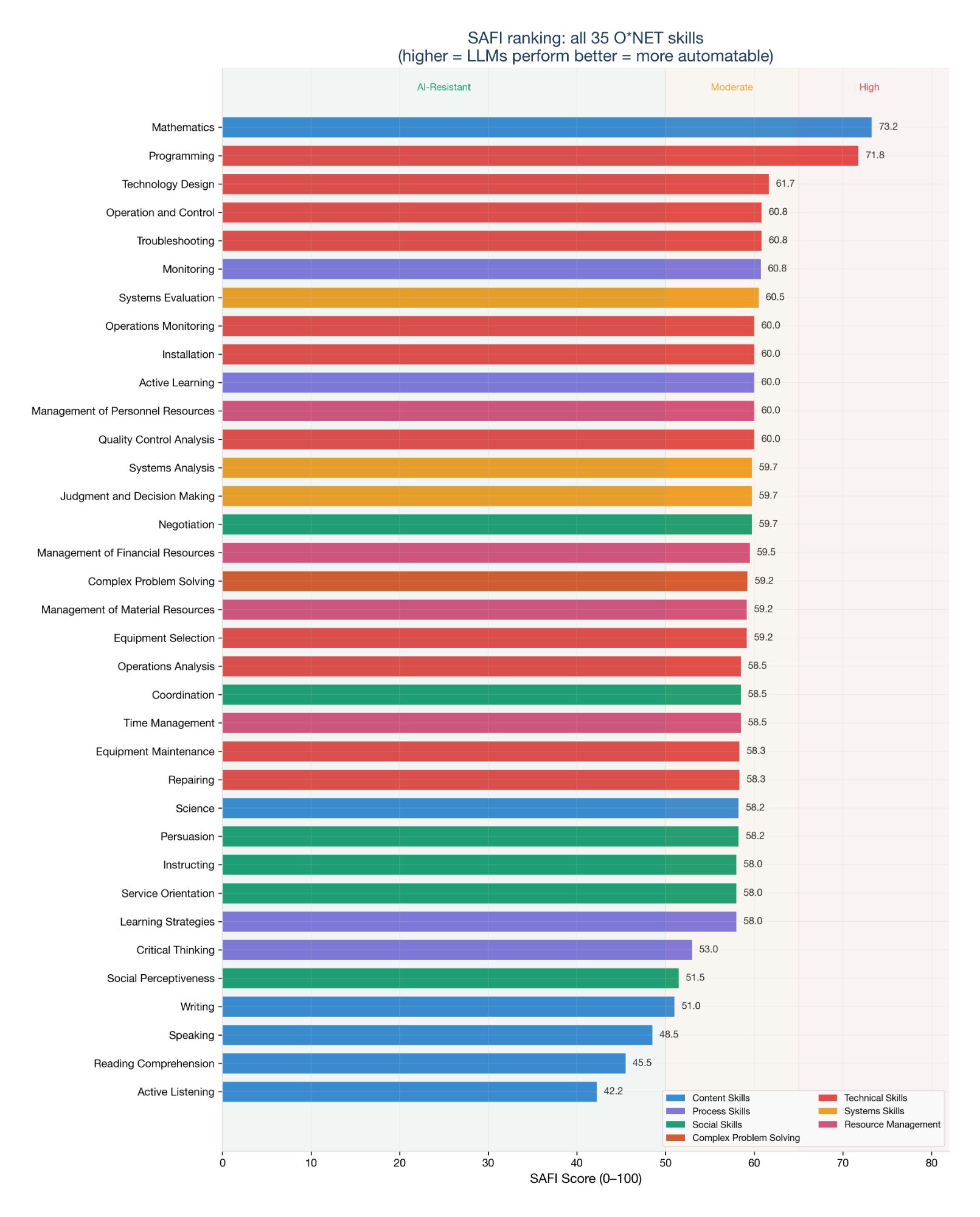}
\caption{SAFI ranking across all 35 O*NET skills. Mathematics and Programming receive the highest text-based automation feasibility scores. Active Listening and Reading Comprehension score lowest. Color indicates skill category; zone annotations indicate relative automation feasibility.}
\label{fig:skill_ranking}
\end{figure}

Figure~\ref{fig:skill_ranking} presents the complete SAFI ranking. Mathematics (73.2) and Programming (71.8) receive substantially higher scores than all other skills, with a clear gap between the top two and the remaining 33 skills (range: 42.2--61.7).

The bottom five skills---Active Listening (42.2), Reading Comprehension (45.5), Speaking (48.5), Writing (51.0), and Social Perceptiveness (51.5)---are all Content or Social skills. These are precisely the skills rated most important across the broadest range of occupations in O*NET, creating a pattern we term the ``capability-demand inversion.''

\begin{table}[h]
\centering
\small
\caption{SAFI scores by skill category. Technical Skills score highest; Content Skills score lowest in text-based evaluation.}
\label{tab:safi_category}
\begin{tabular}{lccr}
\toprule
\textbf{Category} & \textbf{SAFI} & \textbf{$\sigma$} & \textbf{Skills} \\
\midrule
Technical Skills & 62.1 & 0.99 & 11 \\
Systems Skills & 60.0 & 0.67 & 3 \\
Complex Problem Solving & 59.2 & 0.53 & 1 \\
Resource Management & 59.2 & 0.54 & 4 \\
Process Skills & 57.9 & 1.05 & 4 \\
Social Skills & 57.3 & 0.95 & 6 \\
Content Skills & 53.1 & 1.93 & 6 \\
\bottomrule
\end{tabular}
\end{table}

Notably, Content Skills has the highest within-category variance ($\sigma = 1.93$), driven by the extreme range between Mathematics (73.2) and Active Listening (42.2)---both classified as Content Skills in O*NET. This suggests that the Content category contains fundamentally different skill types: structured quantitative reasoning (which LLMs excel at) and nuanced human communication (which LLMs struggle with).

\subsection{SAFI Heatmap: Skill × Model}

\begin{figure}[h!]
\centering
\includegraphics[width=\linewidth]{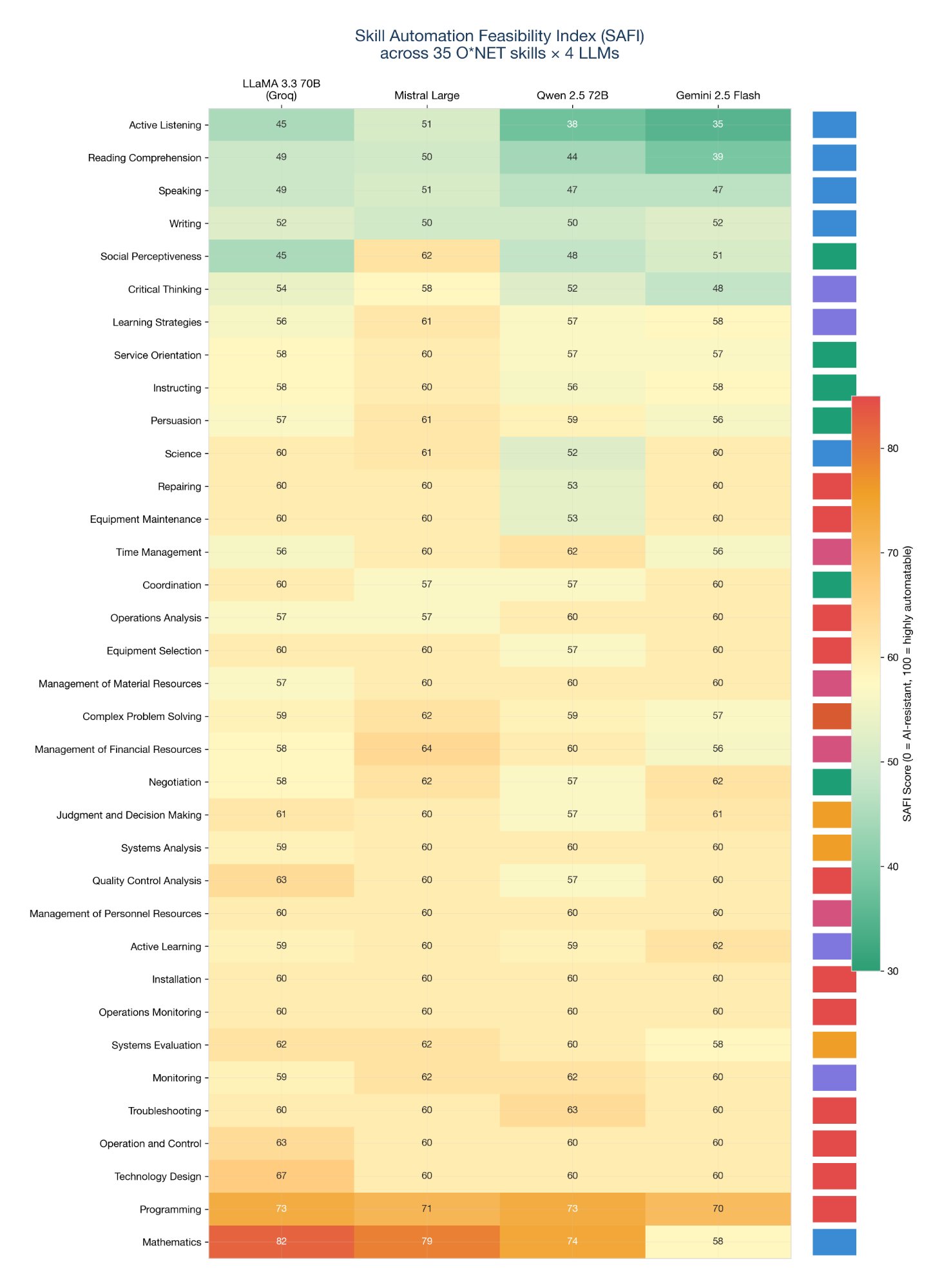}
\caption{Complete SAFI heatmap: 35 skills $\times$ 4 models. Color strip on right indicates skill category. Note the gradient from green (lower SAFI, top) to red (higher SAFI, bottom).}
\label{fig:heatmap}
\end{figure}

Figure~\ref{fig:heatmap} presents the complete SAFI matrix across all 35 skills and 4 models. Several patterns emerge: (1) Mathematics shows the highest cross-model variance, with LLaMA scoring 82 and Gemini scoring 58; (2) the middle band of skills (SAFI 58--62) shows remarkable uniformity across models; (3) Active Listening shows the widest performance gap among low-scoring skills (LLaMA: 45, Gemini: 35).

\subsection{Model Comparison}

\begin{figure}[h!]
\centering
\includegraphics[width=\linewidth]{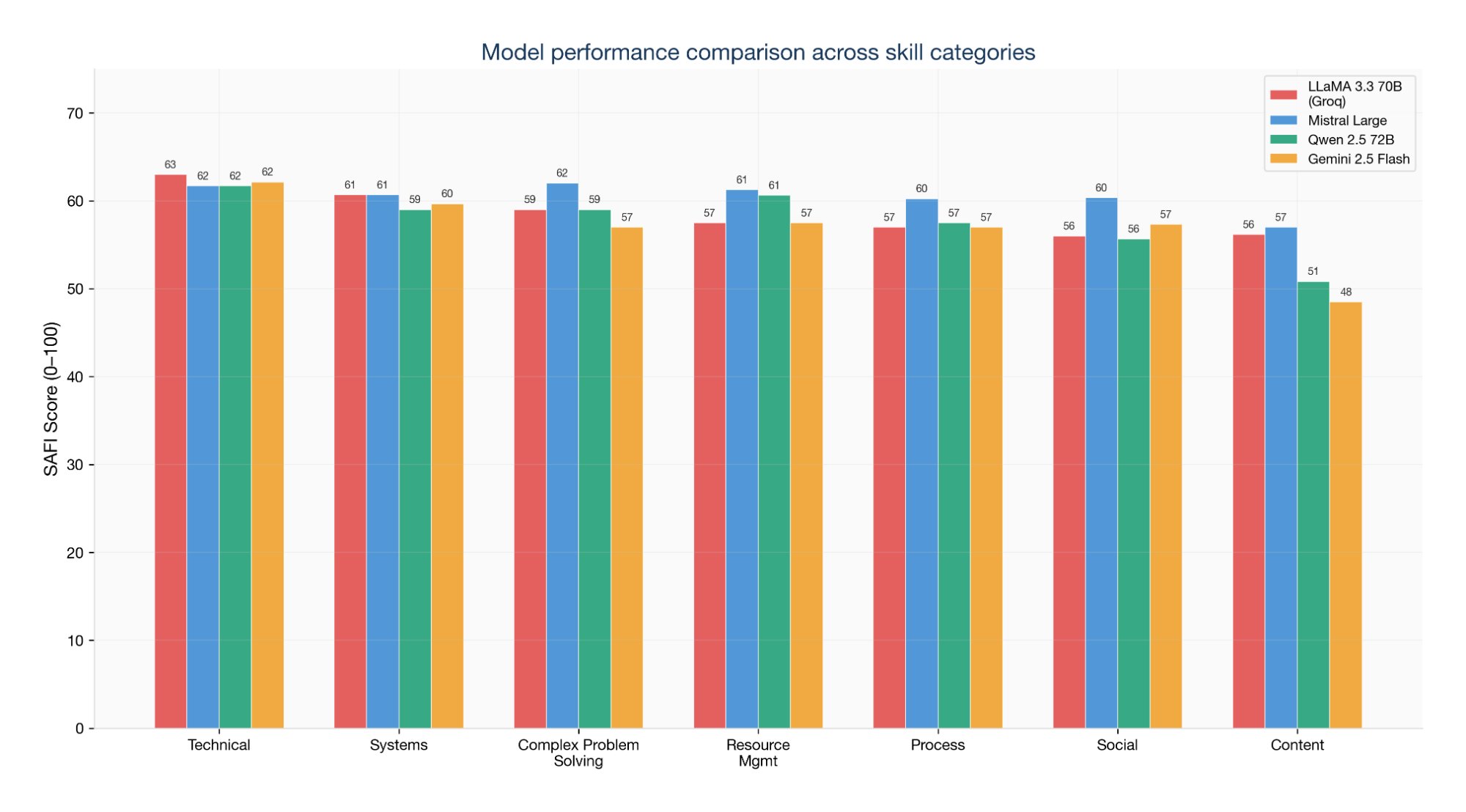}
\caption{SAFI scores by model across skill categories. All four models show similar profiles, with Technical Skills consistently highest and Content Skills lowest.}
\label{fig:model_comparison}
\end{figure}

All four models exhibit remarkably similar performance profiles (Figure~\ref{fig:model_comparison}). Mistral Large achieved the highest overall SAFI (60.0), followed by LLaMA 3.3 70B (58.2), Qwen 2.5 72B (56.7), and Gemini 2.5 Flash (56.4). The narrow 3.6-point spread suggests that \textbf{text-based automation feasibility, as measured here, may be more skill-dependent than model-dependent}.

The most notable inter-model variation occurs in Content Skills, where Mistral Large (57) outperforms Gemini 2.5 Flash (48) by 9 points---the largest gap in any category. This suggests model architecture and training data influence performance on communication-intensive tasks more than on structured reasoning.

\subsection{Difficulty Scaling}

\begin{figure}[h!]
\centering
\includegraphics[width=\linewidth]{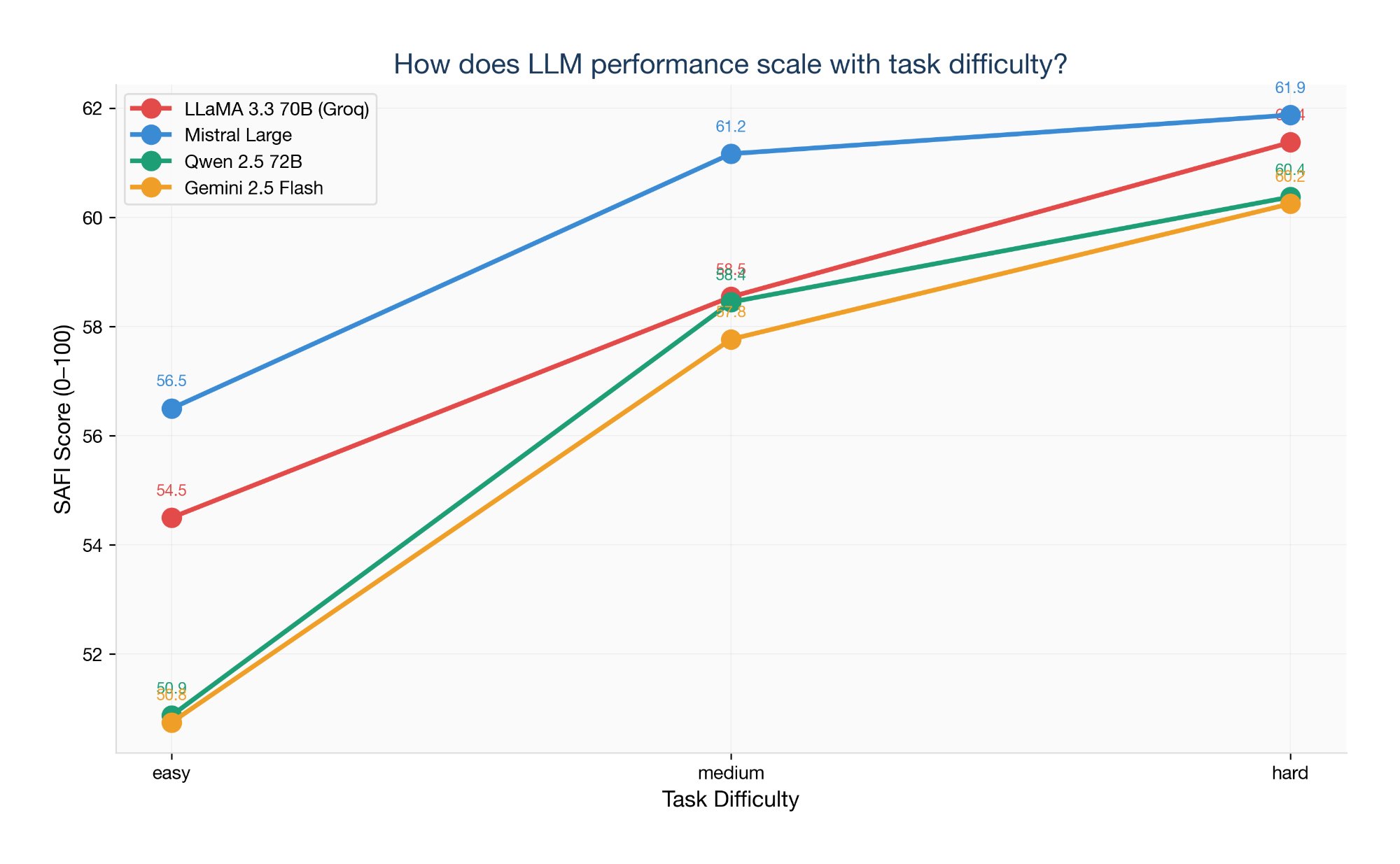}
\caption{SAFI scores by difficulty level and model. All models show higher scores on harder tasks---a scoring artifact of longer, more structured responses rather than genuinely better problem-solving. Mistral Large leads at every difficulty level.}
\label{fig:difficulty}
\end{figure}

All models show increasing SAFI from easy to hard tasks (Figure~\ref{fig:difficulty}). Averaged across models, easy tasks score 53.2, medium tasks 59.0, and hard tasks 61.0. At the per-model level, the pattern is consistent: Mistral Large scores 56.5 / 61.2 / 61.9 across easy / medium / hard; LLaMA 3.3 70B scores 54.5 / 58.5 / 61.4; Qwen 2.5 72B scores 50.9 / 58.4 / 60.4; and Gemini 2.5 Flash scores 50.8 / 57.8 / 60.2. The gap between easy and hard is largest for Qwen (9.5 points) and smallest for Mistral (5.4 points), suggesting Mistral maintains more consistent performance regardless of task complexity. This counterintuitive result---better scores on harder tasks---is a known artifact of our length-sensitive scoring methodology: harder tasks elicit longer, more structured responses that earn higher completeness and reasoning scores. We report this transparently as a limitation of the heuristic scoring approach rather than a finding about model capability.

\subsection{Skill-Exposure Correlations}

\begin{figure}[h!]
\centering
\includegraphics[width=\linewidth]{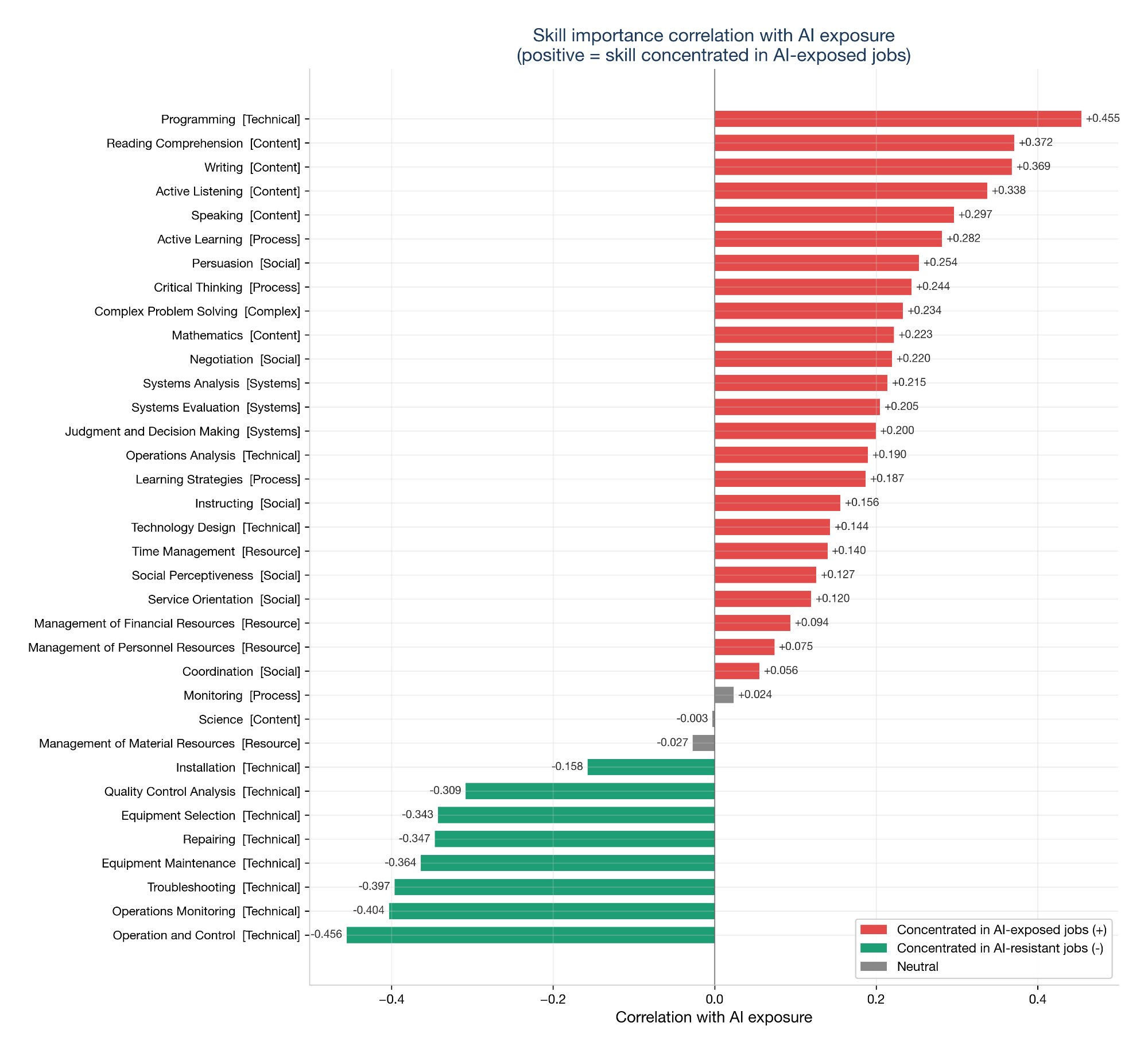}
\caption{Correlation between skill importance and real-world AI exposure across 744 occupations. Programming (+0.455) is most concentrated in AI-exposed jobs; Operation and Control ($-$0.456) is most concentrated in occupations with low AI exposure. The split between cognitive and physical skills is notable.}
\label{fig:exposure_corr}
\end{figure}

Cross-referencing O*NET skill importance ratings with Anthropic Economic Index exposure data across 744 matched occupations yields Pearson correlations for each skill (Figure~\ref{fig:exposure_corr}).

Programming shows the strongest positive correlation (+0.455), meaning it is most important in occupations with high AI exposure. Content skills (Reading Comprehension: +0.372, Writing: +0.369, Active Listening: +0.338) are also concentrated in AI-exposed occupations. In contrast, physical technical skills (Operation and Control: $-$0.456, Operations Monitoring: $-$0.404, Troubleshooting: $-$0.397) concentrate in occupations with low AI exposure.

\subsection{SAFI vs. Real-World Exposure}

\begin{figure}[h!]
\centering
\includegraphics[width=\linewidth]{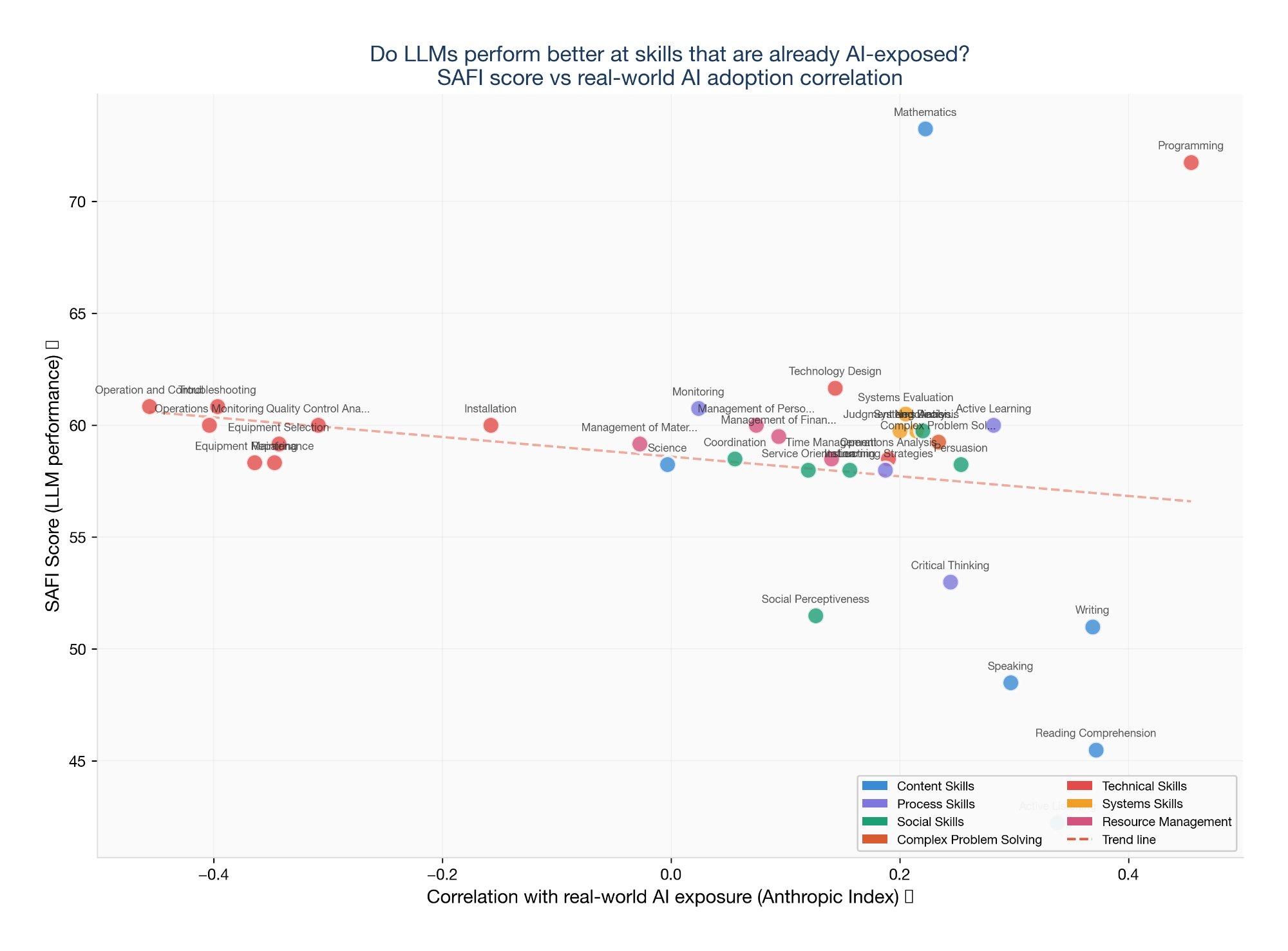}
\caption{SAFI score vs. real-world AI exposure correlation ($r = -0.196$, $p = 0.26$, $n = 35$). The negative trend, while not statistically significant at $\alpha = 0.05$, is consistent with a capability-demand inversion: skills most important in AI-exposed jobs tend to receive lower SAFI scores.}
\label{fig:safi_exposure}
\end{figure}

Figure~\ref{fig:safi_exposure} plots SAFI scores against real-world AI exposure correlations, revealing the capability-demand inversion. The Pearson correlation between SAFI and exposure correlation is $r = -0.196$ ($p = 0.26$, $n = 35$); the Spearman rank correlation is $\rho = -0.300$ ($p = 0.08$). While neither reaches conventional statistical significance---likely due to the small sample of 35 skills---the consistent negative direction across both tests supports the interpretation that skills more important in AI-exposed occupations tend to receive \textit{lower} SAFI scores.

This pattern is consistent with the interpretation that occupations currently adopting AI most heavily may not be the ones most susceptible to skill-level automation. Rather, AI appears to be used predominantly as a \textit{collaborative tool} in roles requiring communication skills where LLMs show lower benchmark performance.

\subsection{Automation vs. Augmentation}

\begin{figure}[h]
\centering
\includegraphics[width=0.85\linewidth]{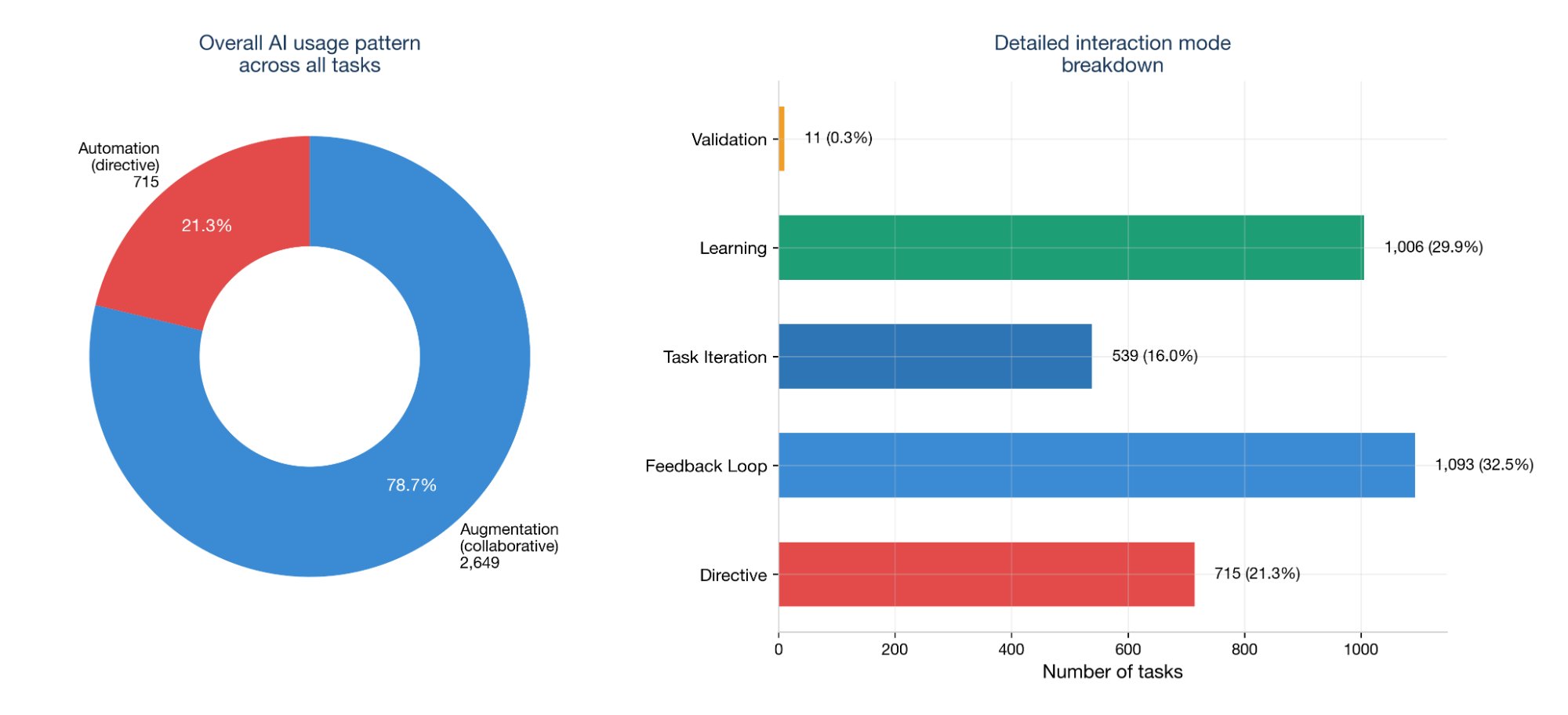}
\caption{Automation vs. augmentation patterns from 3,364 task interactions. Nearly four in five AI interactions are collaborative, not replacement-oriented.}
\label{fig:auto_aug}
\end{figure}

Analysis of 3,364 task-level interaction patterns from the Anthropic Economic Index confirms this interpretation (Figure~\ref{fig:auto_aug}). Of all observed AI-task interactions, 78.7\% represent augmentation (collaborative interaction) versus 21.3\% automation (directive task completion). The dominant augmentation modes are feedback loops (32.5\%), where humans iteratively refine AI outputs, and learning interactions (29.9\%), where AI serves an educational role. Pure directive automation---where the AI completes a task end-to-end without human iteration---accounts for only one in five interactions.

This breakdown is important because it challenges the common narrative of AI as a ``job killer.'' The data suggest that, at present, AI is overwhelmingly used to \textit{enhance} human work rather than \textit{replace} it---a finding consistent with both Anthropic's original analysis \cite{handa2025anthropic} and our SAFI results showing that the skills most involved in AI-exposed occupations are those LLMs score lowest on.

\subsection{Model Skill Profiles}

\begin{figure}[h]
\centering
\includegraphics[width=0.85\linewidth]{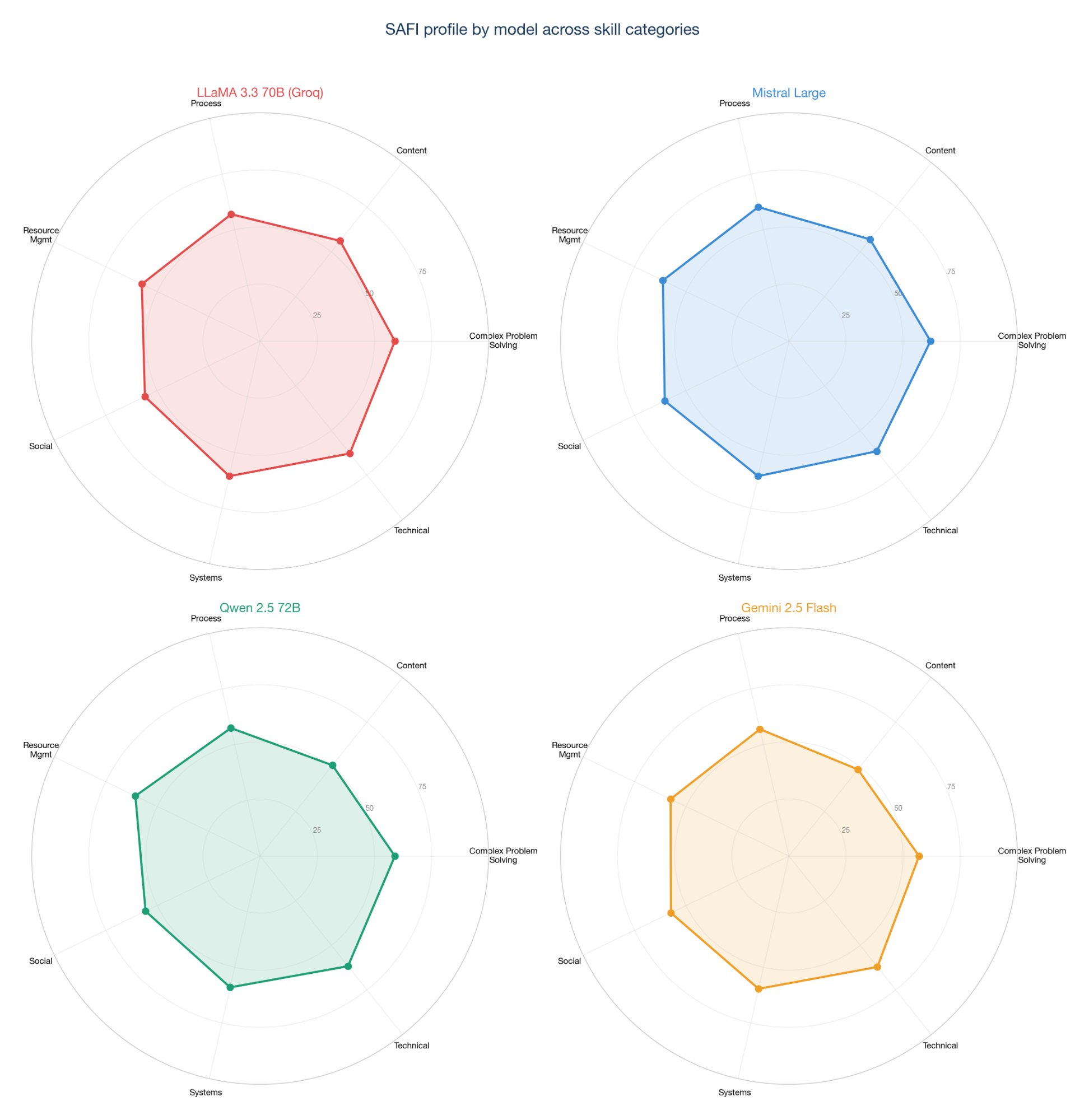}
\caption{SAFI radar profiles for each model across seven skill categories. Mistral Large shows the most balanced capability; all four models converge on similar shapes despite diverse origins.}
\label{fig:radar_models}
\end{figure}

Figure~\ref{fig:radar_models} presents individual SAFI profiles for each model across seven skill categories. Mistral Large shows the most balanced profile, with consistently above-average performance across all categories. LLaMA 3.3 70B shows particular strength in Technical Skills but scores relatively lower on Social Skills. Qwen 2.5 72B and Gemini 2.5 Flash show similar overall shapes despite different origins (China vs. US) and architectures, reinforcing the finding of cross-model convergence.

The convergence of radar shapes across models from three different countries and two licensing regimes (open-source and closed-source) is notable. It suggests that at the 70B+ parameter scale, frontier LLMs develop a shared capability profile for workforce-relevant text tasks---regardless of the specific training pipeline. For workforce planners, this means that skill-level assessments of AI capability are likely to remain relatively stable across the model landscape, at least within the current generation of frontier systems.

% ================================================================
\section{The AI Impact Matrix}

\begin{figure}[h!]
\centering
\includegraphics[width=\linewidth]{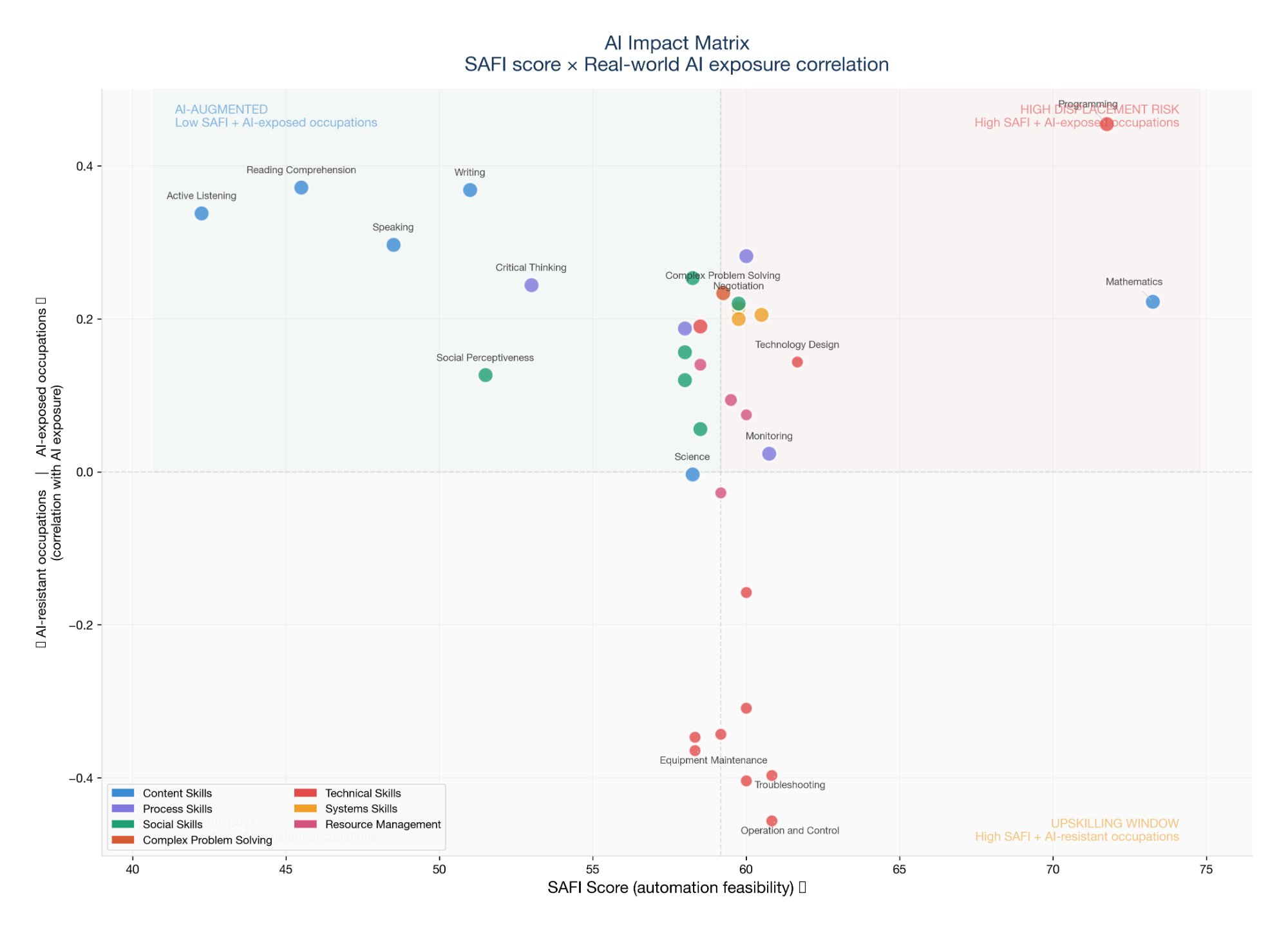}
\caption{The AI Impact Matrix: SAFI score $\times$ real-world AI exposure correlation for all 35 O*NET skills. Each dot represents one skill; color indicates category. Programming sits in the High Displacement Risk quadrant; Content skills cluster in the AI-Augmented quadrant; physical Technical skills fall in the Upskilling Window.}
\label{fig:impact_matrix}
\end{figure}

Synthesizing SAFI benchmarks with real-world AI adoption data, we propose the \textbf{AI Impact Matrix} (Figure~\ref{fig:impact_matrix})---an interpretive framework that positions each skill along two dimensions: text-based automation feasibility (SAFI) and real-world AI exposure correlation. The matrix is intended as a heuristic for structuring workforce planning discussions, not as a direct forecast of displacement outcomes, which depend on many factors beyond LLM text performance.

\textbf{Quadrant I --- Higher Displacement Risk} (High SAFI + Positive AI Exposure). Programming sits here: LLMs score well on its text-based tasks (SAFI: 71.8) and it concentrates in heavily AI-exposed occupations (+0.455). Workers relying primarily on structured programming tasks may face elevated near-term displacement risk, though current evidence points toward AI-augmented development rather than full replacement. In practical terms, this means junior developers and entry-level software engineers---whose work often involves implementing well-specified features, writing boilerplate code, and fixing routine bugs---are more exposed than senior architects who design systems and make judgment calls about tradeoffs. Universities and coding bootcamps may need to shift curricula from syntax fluency toward system design, AI-assisted development workflows, and the ability to evaluate and debug AI-generated code. In our benchmark, LLMs scored highest on self-contained coding tasks (writing functions, implementing algorithms, generating SQL queries) and lowest on tasks requiring multi-file architectural reasoning or ambiguous specification interpretation---reinforcing the distinction between automatable routine coding and resilient system-level design work.

\textbf{Quadrant II --- AI-Augmented} (Low SAFI + Positive AI Exposure). Content skills (Reading, Writing, Speaking, Active Listening) and some Social skills (Persuasion, Negotiation) cluster here. AI is used heavily in occupations requiring these skills, but as a \textit{collaborative tool}---LLMs cannot replicate the nuanced human communication these skills demand. The implications are concrete: customer support representatives are using AI to draft responses faster, but the empathy, de-escalation, and contextual judgment that define excellent service remain human. Financial analysts use LLMs to summarize reports and generate first drafts, but the critical reading, skeptical evaluation, and client-facing communication that drive investment decisions stay with the analyst. In education, teachers may use AI to generate lesson materials, but active listening to a struggling student and adapting instruction in real time are irreplaceable. For entry-level white-collar workers---administrative assistants, junior analysts, content coordinators---the shift is not from employment to unemployment, but from routine execution to AI-augmented productivity, where the workers who thrive will be those who learn to direct, evaluate, and refine AI outputs effectively.

\textbf{Quadrant III --- Upskilling Window} (Moderate/High SAFI + Negative AI Exposure). Physical Technical skills (Equipment Maintenance, Troubleshooting, Operation and Control) appear here. LLMs can \textit{discuss} these skills abstractly (moderate SAFI) but the occupations requiring them are not AI-exposed because the skills demand physical presence. As AI-assisted diagnostics, predictive maintenance platforms, and remote monitoring advance, this quadrant represents a \textit{window} for proactive upskilling. HVAC technicians, manufacturing operators, and field maintenance workers are currently insulated from AI disruption---but as sensor data and AI diagnostic tools enter their workflows, those who can combine hands-on expertise with data literacy will be best positioned. Trade schools and vocational programs have an opportunity to integrate AI-assisted troubleshooting into their curricula now, before the transition accelerates.

\textbf{Quadrant IV --- Lower Displacement Risk} (Low SAFI + Negative/Neutral AI Exposure). Skills requiring embodied human judgment in occupations with low current AI exposure fall here, suggesting relatively lower near-term disruption from text-based AI systems.

% ================================================================
\section{Discussion}

\subsection{The Capability-Demand Inversion}

Our most notable finding is that the skills most concentrated in AI-exposed occupations are not the skills LLMs score highest on in our text-based benchmark. This ``capability-demand inversion'' has important implications for workforce planning. It is consistent with the interpretation that the current AI adoption wave is driven by \textit{augmentation demand}---workers in communication-heavy roles using AI as a productivity tool---rather than by \textit{automation capability}---AI replacing the core skills of those roles.

If this pattern holds, the most immediate workforce challenge may not be mass displacement of communication-intensive roles, but rather a gradual shift in what those roles require: from pure execution to AI-augmented execution, demanding new meta-skills like prompt engineering, AI output evaluation, and human-AI workflow design. Consider the paralegal who now uses AI to draft legal summaries but must still catch hallucinated case citations, or the marketing analyst who generates campaign copy with AI but must ensure it resonates with the target audience's cultural context. The skill itself is not automated; the \textit{workflow around the skill} is restructured.

\subsection{Model Convergence}

The narrow 3.6-point SAFI spread across four diverse models (US, France, China; open-source and closed-source) suggests that current frontier LLMs may have converged to a similar capability profile for text-based workforce skill tasks. If confirmed by broader evaluations, this would mean workforce planning need not track individual model releases---the skill-level performance profile appears relatively stable across the current frontier.

\subsection{The Content Skills Paradox}

Content Skills simultaneously contain the skill with the highest SAFI score (Mathematics: 73.2) and the lowest (Active Listening: 42.2). This 31-point within-category spread---the largest of any category---highlights a distinction between \textit{structured content processing} (mathematical reasoning, where LLMs score well) and \textit{unstructured content understanding} (active listening, which in occupational practice involves emotional intelligence, contextual inference, and nonverbal cue interpretation---dimensions that text-based models are not designed to capture).

\subsection{Industry Perspectives and Policy Implications}

The debate among industry leaders illustrates precisely why empirical skill-level data matters. Dimon's position---that displacement is happening now and society needs phased responses including ``retraining, relocation, and income assistance'' \cite{dimon2026feb}---implies a need for granular knowledge of \textit{which} skills are affected. Solomon's counterposition---that the economy is ``incredibly broad and nimble'' enough to absorb displaced workers \cite{solomon2026}---still requires understanding where the absorption will occur. Our SAFI index and AI Impact Matrix may help structure both perspectives.

Solomon offered a vivid illustration of AI's productivity impact in financial services: at Goldman Sachs, AI can now draft 95\% of an IPO prospectus (S-1 filing) in minutes---a task that previously required a six-person team working for two weeks \cite{solomon2025ipo}. Yet as Solomon noted, ``the last 5\% now matters because the rest is now a commodity.'' This precisely mirrors our Quadrant II finding: the structured, text-amenable components of financial analysis (high SAFI) are being automated, while the judgment, client communication, and regulatory interpretation (low SAFI) become more valuable.

While our framework cannot directly predict displacement outcomes, it may help differentiate policy responses: skills in Quadrant I (Higher Displacement Risk) warrant attention for transition planning; skills in Quadrant II (AI-Augmented) suggest training in AI collaboration tools may be beneficial; skills in Quadrant III (Upskilling Window) may offer time for proactive preparation. For governments, this means workforce retraining programs should not treat ``AI exposure'' as monolithic---a data entry clerk (Quadrant I) needs a fundamentally different intervention than a customer service representative (Quadrant II) or an equipment technician (Quadrant III). For corporations, internal training budgets may be better allocated toward AI collaboration skills for communication-heavy roles than toward wholesale role elimination. And for individual workers, the message is nuanced: the question is less ``will AI take my job?'' and more ``how will AI change what my job requires?''

\subsection{Recommendations}

Based on our findings, we offer the following actionable recommendations, organized by stakeholder.

\textbf{For policymakers and governments:} Workforce retraining programs should be differentiated by skill quadrant, not treated as one-size-fits-all ``AI readiness'' initiatives. Workers in Quadrant I occupations (e.g., data entry clerks, junior programmers) need funded transition pathways to adjacent roles---our SAFI scores, combined with O*NET skill-adjacency data, could help identify the shortest reskilling routes. Workers in Quadrant II roles (e.g., analysts, customer support, educators) need subsidized training in AI collaboration tools---prompt engineering, output verification, and human-AI workflow design. Workers in Quadrant III occupations (e.g., HVAC technicians, manufacturing operators) have a narrow but real window for proactive upskilling before AI diagnostics reshape their fields. Community colleges and vocational programs should integrate AI-assisted troubleshooting and data literacy modules now, while the window remains open.

\textbf{For corporations:} Rather than framing AI adoption as headcount reduction, our data supports a \textit{redeployment model} consistent with JPMorgan's approach \cite{dimon2026feb}. Internal training investments should prioritize teaching existing employees to work \textit{with} AI---evaluating outputs, catching errors, maintaining quality---rather than replacing them. The 78.7\% augmentation rate suggests that most current AI usage already follows this pattern; formalizing it through structured training programs is the logical next step.

\textbf{For educational institutions:} University curricula---particularly in computer science, business, and communication---should evolve to emphasize the skills that sit at the intersection of high labor market demand and low AI capability: critical evaluation of AI-generated content, complex interpersonal communication, ethical judgment under uncertainty, and the ability to design workflows that combine human strengths with AI efficiency. The era of teaching skills in isolation from their AI context is ending.

\textbf{For individual workers:} Identify where your primary skills sit on the AI Impact Matrix. If you rely heavily on Quadrant I skills (structured programming, data processing), invest in system-level thinking, architecture, and AI-augmented development practices. If your work centers on Quadrant II skills (writing, analysis, communication), learn to use AI as a force multiplier---the workers who thrive will not be those who resist AI, but those who integrate it most effectively into their craft.

\subsection{Limitations}

Several important limitations should be noted. First, and most fundamentally, \textbf{SAFI measures LLM performance on text-based task representations, not the full occupational execution of skills.} Many O*NET skills---particularly Social and Technical skills---involve physical, embodied, real-time, or interpersonal dimensions that cannot be captured through text prompts. A high SAFI score for a skill like ``Operation and Control'' reflects the model's ability to \textit{discuss} the skill's cognitive components, not its ability to physically operate machinery. Readers should interpret SAFI as an upper-bound proxy for the text-amenable component of each skill, not as a direct measure of occupational automation feasibility.

Second, our heuristic scoring methodology, while avoiding LLM-as-judge bias, relies on surface-level response features (length, structural markers, reasoning keywords) that may not fully capture response quality---particularly for nuanced Social Skills tasks where expert human evaluation would be more appropriate. The scoring does not verify factual correctness of responses.

Third, 263 tasks across 35 skills, while covering the full O*NET taxonomy, represent a limited sample of the vast space of possible skill applications. Skills with fewer tasks (3 tasks for some Technical skills vs. 10 for Content skills) have less statistical power.

Fourth, Anthropic Economic Index data reflects Claude usage specifically and may not generalize to all AI platforms or to non-English-speaking labor markets.

Fifth, our study is cross-sectional; SAFI scores will change as models improve, and the AI Impact Matrix reflects a snapshot of current capabilities rather than a forecast.

Sixth, we do not incorporate BLS employment data, which would allow weighting by workforce size and estimating the number of workers affected; this is planned for future work.

Finally, the AI Impact Matrix is an \textit{interpretive framework} for organizing findings, not a predictive model. Actual displacement outcomes depend on many factors beyond LLM text performance, including organizational adoption decisions, regulatory environments, economic conditions, and the pace of complementary technology development.

% ================================================================
\section{Future Work}

We identify four priority directions for extending this research:

\textbf{(1) Reskilling Transition Pathway Maps.} Using SAFI scores combined with O*NET's skill-adjacency data (which tracks how similar skills are across occupations), we plan to construct shortest-path transition maps that recommend specific career moves for workers in high-displacement-risk occupations. For example: a data entry clerk (Quadrant I, high displacement risk) shares skill overlap with administrative coordinators and project assistants (Quadrant II, AI-augmented)---quantifying these transitions would make retraining programs more targeted and efficient.

\textbf{(2) Longitudinal SAFI Tracking.} As new models are released (GPT-5, Claude 4, Gemini 3, LLaMA 4), re-running our benchmark would measure how fast the automation frontier is advancing per skill. If Active Listening's SAFI jumps from 42 to 65 in 18 months, that changes the policy calculus significantly. We plan to establish a semi-annual benchmarking cadence.

\textbf{(3) AI-Emergent Skill Taxonomy.} O*NET's 35 skills were designed before LLMs existed. New skills have emerged---prompt engineering, AI output evaluation, multi-agent orchestration, human-AI workflow design---that are not captured in the current taxonomy. Cataloging and benchmarking these emergent skills would complete the picture of the evolving labor market.

\textbf{(4) BLS Employment Integration.} Incorporating Bureau of Labor Statistics employment data would allow us to weight SAFI scores by the number of workers affected, transforming skill-level insights into workforce-level impact estimates (e.g., ``Programming automation feasibility affects approximately 1.8 million U.S. workers'').

% ================================================================
\section{Conclusion}

The prevailing narrative about AI and work is binary: either AI will take your job, or it won't. Our data suggests a third possibility that is both more nuanced and more urgent.

The capability-demand inversion reveals that AI is not advancing uniformly across the skill landscape. It is exceptionally strong at structured, rule-bound reasoning---mathematics, programming, systems analysis---and measurably weak at the unstructured, deeply human skills that the labor market values most: listening, reading, speaking, writing, social perception. These are not peripheral skills. They are the connective tissue of the modern economy, the skills that make organizations function, clients trust, and teams collaborate.

This asymmetry means that the most significant economic impact of AI in the near term is not displacement but \textit{restructuring}. The job stays. The workflow changes. The human becomes responsible not for the first draft but for the last mile---the judgment, the nuance, the contextual awareness that a language model trained on internet text cannot access. This is augmentation in a precise, measurable sense: 78.7\% of real-world AI interactions already follow this pattern.

But this finding is not cause for complacency. It is cause for urgency. The capability-demand inversion holds \textit{today}. It holds for \textit{this generation} of models, benchmarked at the 70B-parameter frontier in early 2026. There is no guarantee it will hold in 2028. If future models close the gap on communication and social skills---as multimodal architectures, real-time audio models, and embodied AI agents suggest they might---the augmentation-dominant equilibrium could shift rapidly toward automation.

The window for proactive preparation is open. It will not stay open indefinitely.

We release our complete task battery, all 1,052 model responses, SAFI scores, and analysis code at \url{https://github.com/rudrajadhav/ai-skills-shift}---not as a finished answer, but as a foundation. The question of which skills AI can and cannot perform is not static. It requires continuous, empirical measurement. We hope this work contributes to a culture of evidence over speculation in a debate where the stakes are measured in livelihoods.

% ================================================================
\section*{Data Availability}
All datasets used in this study are publicly available: O*NET Database (v30.2) from \url{onetcenter.org}, Anthropic Economic Index from \url{huggingface.co/datasets/Anthropic/EconomicIndex}. Our benchmark task battery, model responses, and SAFI scores are released under MIT license.

\section*{Acknowledgments}
We thank the developers of the O*NET database, Anthropic for open-sourcing the Economic Index data, and the teams behind LLaMA (Meta), Mistral (Mistral AI), Qwen (Alibaba), and Gemini (Google) for providing API access. Special thanks to the Groq and HuggingFace teams for free inference infrastructure that made this research possible without institutional funding.

% ================================================================
% ================================================================

\end{document}